\documentclass[sigconf]{acmart}
\usepackage{multicol, multirow}
\usepackage{enumitem}
\usepackage{tcolorbox}
\tcbuselibrary{listings,breakable}
\usepackage{comment}
\AtBeginDocument{%
  }

\setcopyright{none}
\copyrightyear{2025}
\acmYear{2025}
\acmDOI{}
\acmConference[FinAI@CIKM '25]{CIKM 2025 Workshop on Advances in Financial AI: Innovations, Risk, and Responsibility in the Era of LLMs (Non-archival)}{November 14, 2025}{Seoul, Korea}
\acmISBN{}

\begin{document}

\title{FinNuE: Exposing the Risks of Using BERTScore for Numerical Semantic Evaluation in Finance}


\author{Yu-Shiang Huang}
\authornote{These authors contributed equally to this work.}
\affiliation{%
  \institution{Data Science Degree Program, National Taiwan University and Academia Sinica}
  \city{Taipei}
  \country{Taiwan}}
\email{F09946004@ntu.edu.tw}

\author{Yun-Yu Lee}
\authornotemark[1]
\affiliation{%
  \institution{Department of Computer Science, National Yang Ming Chiao Tung University}
  \city{Hsinchu}
  \country{Taiwan}}
  \email{raelee.cs12@nycu.edu.tw}

\author{Tzu-Hsin Chou}
\authornotemark[1]
\affiliation{%
  \institution{Department of Information Management, National Taiwan University}
  \city{Taipei}
  \country{Taiwan}}
  \email{dorachou0609@gmail.com}

\author{Che Lin}
\affiliation{%
  \institution{Department of Electrical Engineering, National Taiwan University}
  \city{Taipei}
  \country{Taiwan}}
  \email{chelin@ntu.edu.tw}

\author{Chuan-Ju Wang}
\affiliation{%
  \institution{Research Center for Information Technology Innovation, Academia Sinica}
  \city{Taipei}
  \country{Taiwan}}
\email{cjwang@citi.sinica.edu.tw}

\renewcommand{\shortauthors}{Huang et al.}

\begin{abstract}
BERTScore has become a widely adopted metric for evaluating semantic similarity between natural language sentences. 
However, we identify a critical limitation: BERTScore exhibits low sensitivity to numerical variation, a significant weakness in finance where numerical precision directly affects meaning (e.g., distinguishing a 2\% gain from a 20\% loss).
We introduce \textbf{FinNuE}, a diagnostic dataset constructed with controlled numerical perturbations across earnings calls, regulatory filings, social media, and news articles. 
Using FinNuE, demonstrate that BERTScore fails to distinguish semantically critical numerical differences, often assigning high similarity scores to financially divergent text pairs. Our findings reveal fundamental limitations of embedding-based metrics for finance and motivate numerically-aware evaluation frameworks for financial NLP.
\end{abstract}



\received{21 September 2025}     
\received[revised]{30 September 2025}   
\received[accepted]{31 October 2025} 

\maketitle

\section{Introduction}

Evaluating natural language generation systems requires measuring how well the system outputs match reference texts. 
Lexical overlap metrics such as BLEU~\cite{papineni-etal-2002-bleu} and ROUGE~\cite{lin-2004-rouge} dominated early work but fail to capture semantic equivalence~\cite{reiter-2018-structured, novikova-etal-2017-need}. 
Embedding-based metrics, particularly BERTScore \cite{Zhang2020BERTScore}, have emerged as the preferred alternative by leveraging contextual embeddings from pre-trained Transformers to compute semantic similarity.
BERTScore has achieved remarkable adoption across natural language generation (NLG) tasks,  including widespread use in financial NLP benchmarks~\cite{chen2024fingen, NEURIPS2024_finben, son2023finreason, wang-etal-2025-finnlp, cao-etal-2025-capybara-finben, mukherjee-etal-2022-ectsum}.
However, we identify a critical limitation: BERTScore exhibits low sensitivity to numerical variation, a significant weakness in financial domains where numerical precision fundamentally determines semantic meaning.
Consider two sentences: ``\emph{The company reported a 2\% revenue increase}'' and ``\emph{The company reported a 20\% revenue increase.}'' While semantically distinct with vastly different financial implications, BERTScore assigns these a similarity score of 0.97, failing to capture meaningful numerical differences.

To systematically investigate this limitation, we introduce \textbf{Fin-NuE} (\textbf{Fin}ancial \textbf{Nu}merical \textbf{E}valuation), a diagnostic dataset designed to test numerical sensitivity in evaluation metrics.
FinNuE contains 19,578 sentence pairs sourced from earnings calls, regulatory filings, financial news, and social media, where each pair differs only in numerical values (e.g., percentages, dollar amounts, dates) while maintaining grammatical structure. This controlled design isolates numerical understanding from other semantic factors.
We evaluate BERTScore under two complementary protocols. 
First, in anchor-based evaluation, we assess whether BERTScore assigns lower similarity to sentences with larger numerical deviations from a reference (e.g., 2\% vs. 5\% vs. 20\% should show decreasing similarity).
Second, in cross-pair evaluation, we examine whether BERTScore preserves correct similarity rankings when comparing sentence pairs from different contexts (e.g., the similarity between "Revenue grows 30\%" and "Revenue grows 32\%" should be higher than the one between "Expense accounts for 30\%" and "Expense accounts for 60\%").
Our findings reveal that BERTScore shows limited sensitivity to numerical magnitude, often assigning similar scores to sentences with substantially different numerical values (e.g., 2\% vs. 200\%). 
These results suggest that numerically-aware evaluation metrics that incorporate structured understanding of quantities alongside semantic embeddings may be necessary for financial NLP applications.

\begin{figure*}[t]
    \centering
    \includegraphics[width=0.8\linewidth]{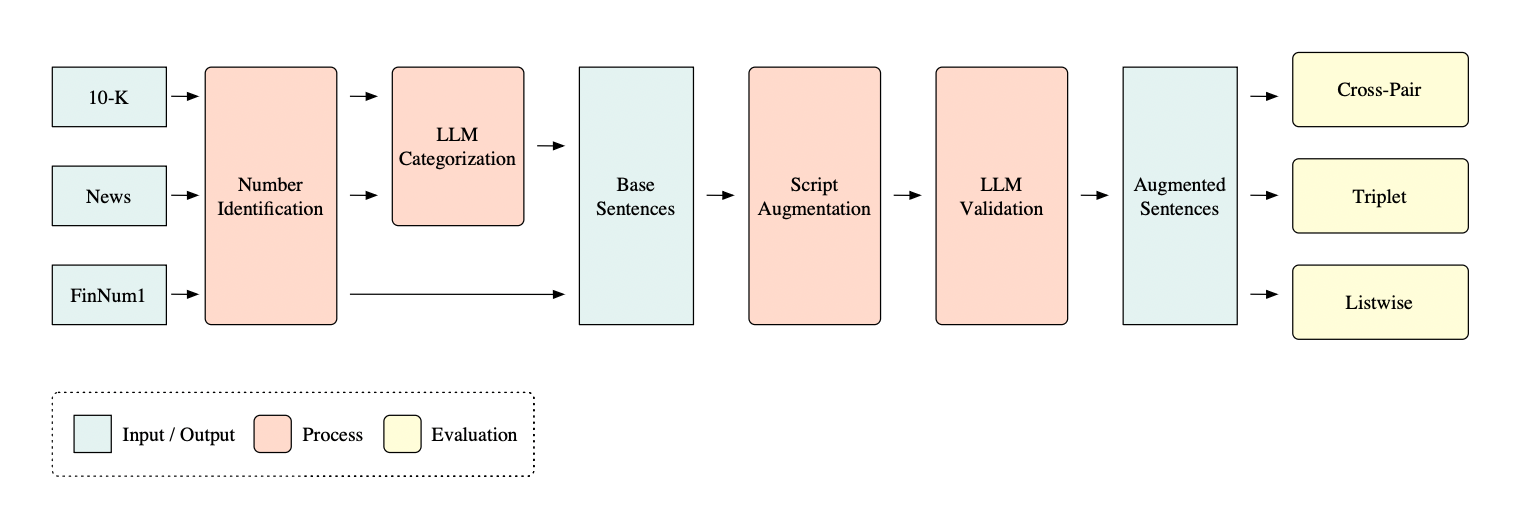}
    \caption{Overview of the FinNuE dataset construction pipeline.}
    \label{fig:pipeline}
\end{figure*}

\section{Motivation}
\label{sec:motivation}

BERTScore compares sentences by aggregating token-level embedding similarities. 
Given a sentence pair \( (s, \hat{s}) \), each sentence is tokenized into a sequence:
\begin{align} \label{eq:tokenizer}
s &= (w_1, \ldots, w_m) = \mathcal{T}(s), \quad \hat{s} = (\hat{w}_1, \ldots, \hat{w}_n) = \mathcal{T}(\hat{s}), 
\end{align}
where \(\mathcal{T}\) denotes the tokenizer.
Each token is encoded by a pre-trained Transformer \(f_\theta\), producing contextualized embeddings
$h^s_i = f_\theta(w_i)$, and $h^{\hat{s}}_j = f_\theta(\hat{w}_j)$ in
$\mathbb{R}^d$.
For each token \(w_i\), BERTScore computes its maximum similarity to any token in \(\hat{s}\):
\begin{align} \label{eq:token-align}
\text{sim}(w_i, \hat{s}) = \max_{j} \cos(h^s_i, h^{\hat{s}}_j).
\end{align}
The sentence-level BERTScore aggregates these similarities with optional importance weights \(\delta_i\) (e.g., IDF):
\begin{align}
\texttt{BERTScore}(s,\hat{s}) = 
\frac{\sum_{i=1}^m \delta_i \cdot \text{sim}(w_i, \hat{s})}{\sum_{i=1}^m \delta_i}.
\end{align}

This formulation, however, treats numerical tokens no differently from other words.
Consider three sentences with identical structure but varying numerical values:
\begin{align*}
s_1 &: \; \text{``Revenue increased by 3.56\%.''}, \\ 
s_2 &: \; \text{``Revenue increased by 4\%.''}, \\  
s_3 &: \; \text{``Revenue increased by 40\%.''}.
\end{align*}
We would expect
$\texttt{BERTScore}(s_1, s_2) > \texttt{BERTScore}(s_2, s_3)$, as the numerical difference between $s_1$ and $s_2$
(0.44 percentage points) is far smaller than between $s_2$ and $s_3$ (36 percentage points).
Using the widely adopted HuggingFace implementation,\footnote{We report \texttt{BERTScore-F1} from \url{https://huggingface.co/spaces/evaluate-metric/bertscore}.}, we observe the opposite: $
\texttt{BERTScore}(s_1, s_2) = 0.9639  <  \texttt{BERTScore}(s_2, s_3) = 0.9764.$
This counterintuitive result reveals a fundamental limitation: BERTScore's token-level matching cannot distinguish between semantically critical numerical differences.
While existing semantic similarity datasets~\cite{cer-etal-STSB, liu-etal-2024-finSTS} rely on subjective human judgments, numerical distinctions in financial text offer an objective and interpretable signal for evaluating metric quality. 
To systematically investigate this limitation, we construct \textbf{FinNuE}, a diagnostic dataset designed to test whether evaluation metrics capture semantically meaningful numerical differences across diverse financial contexts.

\section{FinNuE Dataset}

FinNuE is constructed to isolate numerical understanding from other semantic factors through controlled perturbations.
 Figure~\ref{fig:pipeline} illustrates our construction pipeline: we (1) extract sentences with numerical content from diverse financial sources, (2) systematically augment numerical tokens while preserving grammatical structure, (3) apply quality filters to ensure semantic validity, and (4) design complementary evaluation protocols to test metric sensitivity to numerical variation.

\subsection{Data Sources}
\textbf{FinNuE} draws from three complementary financial texts to ensure broad domain coverage. Social media is represented by the {FinNum1} dataset~\cite{finnum}. 
Financial news is sourced from {Financial Phrasebank}~\cite{FinPhraseBank}, from which we retain sentences with at least 50\% annotator agreement to ensure quality. 
Regulatory disclosures are sampled from {10-K filings} (fiscal years 2023–2024) of 55 publicly traded companies: the five largest by market capitalization in each of the eleven GICS sectors.
This stratified sampling ensures representation across industries, including technology, healthcare, finance, and energy.
Together, these sources yield 9,227 sentences containing diverse numerical expressions—percentages, dollar amounts, dates, and financial metrics—that serve as the foundation for controlled augmentation.

\begin{table}[ht]
\centering\small
\caption{The seven rule-based augmentation transformations applied in FinNuE with examples. $k$ denotes the number of variants generated per evaluation unit.}
\vspace{-0.4cm}
\begin{tabular}{l l l c}
\toprule
\textbf{Augmentation} & \textbf{Target Number} & \textbf{Variant} & \textbf{$k$} \\
\midrule
\texttt{Date Shift}  & Sep. 28, 2025   & Sep. 28, 2029  & 9 \\
\texttt{Duration Convert}   & 1 week & 7 days; 1 month & 9 \\
\texttt{Extra Decimal}      & 3.5    & 3.56   & 9 \\
\texttt{Fractional Shift}   & 0.25    & 0.37; 0.13    & 9 \\
\texttt{Scale Change}       & 1,000  & 10,000 & 9 \\
\texttt{Million to Billion} & 110 million & 0.11 billion & 1 \\
\texttt{Last Digit Edit}    & 110    & 1100; 11 & 2 \\
\bottomrule
\end{tabular}
\label{tab:rule-based-augmentation}
\end{table}

\subsection{Numerical Augmentation}
\subsubsection{Numerical Extraction and Categorization}
We identify every numeral as a potential augmentation target and categorize it according to FinNum1's taxonomy~\cite{finnum}, which distinguishes seven types: temporal expressions, monetary values, percentages, quantities, product numbers, indicators, and option values.
For Financial Phrasebank and 10-K sentences, categories and subcategories are automatically assigned using GPT-4o\footnote{ Prompts provided in Appendix~\ref{app:prompts}.}. 
Each sentence containing at least one numeral becomes a \emph{base sentence} with target number $t$, where $\nu(t)$ denotes the numerical value (e.g., $\nu(t) = 15$
for ``15\%'' or ``15M'').

\subsubsection{Augmentation Strategy}
To systematically test numerical sensitivity, we apply two complementary augmentation approaches that modify target numbers while preserving grammatical structure.
Since base sentences may contain multiple target numbers, and each can be modified through different augmentation methods, we define an \emph{evaluation unit} as a tuple:
$(s, t, \alpha)$,
where \(s\) is a base sentence, \(t\) is a target number in \(s\), and \(\alpha\) specifies the augmentation types. 
Each evaluation unit produces $k$ augmented variants:
$ (s, t, \alpha) \;\mapsto\; \{ \tilde{s}_1, \tilde{s}_2, \ldots, \tilde{s}_k \}, $
where \(\tilde{s}_i\) results from applying augmentation \(\alpha\) to \(t\) in \(s\). 
The two augmentation types are: 

\noindent\textbf{Random augmentation} perturbs each target number within specific bounds to test whether metrics are sensitive to \emph{magnitude}. 
For each base sentence with a target number, we generate $k=9$ variants by randomly increasing or decreasing $\nu(t)$ (e.g., shifting percentages by ±10\%, ±25\%, ±50\%, ±100\%). 
This produces a spectrum of numerical deviations from the original value. 
Detailed augmentation ranges for each category appear in 
Appendix~\ref{app:random-augmentation}.
    
\noindent\textbf{Rule-based augmentation} applies category-specific transformations that test fine-grained numerical understanding: \texttt{Date Shift} , \texttt{Duration Convert}, \texttt{Extra Decimal}, \texttt{Fractional Shift}, \texttt{Scale Change}, \texttt{Million to Billion}, and \texttt{Last Digit Edit}. 
Depending on the rule, $k\in\{1,2,9\}$ variants are generated for each applicable base sentence with a target number.
Examples appear in Table~\ref{tab:rule-based-augmentation}.
In total, random augmentation generates 19,365 evaluation units, while rule-based augmentation produces 27,857 units. 


\subsection{Quality Assurance}
To ensure augmented sentences remain both grammatically valid and semantically plausible, we employ {GPT-4o-mini} as an automatic validator.\footnote{Prompts provided in Appendix~\ref{app:prompts}.}
 The model evaluates each variant along two dimensions: (1) {\bf temporal validity} (whether dates and times follow calendar conventions, e.g., no ``February 31st'') and (2) {\bf numerical plausibility} (whether shifted values remain contextually reasonable, e.g., not ``a 500\% profit margin''). 
GPT-4o-mini assigns each variant a validity score in \([0.0, 1.0]\).

We discard evaluation units where more than 3 variants (out of 9 for random, 1–9 for rule-based) receive scores below 0.5, indicating systematic issues with the augmentation. 
This filtering retains {75.66\% of random augmentation units (14,651 valid units) and 85.17\% of rule-based units (23,730 valid units). 
Detailed statistics appear in Table~\ref{tab:validated-dataset-stats}.  

To validate this automated process, we manually inspected 50 randomly sampled evaluation units. 
Human judges agreed with GPT-4o-mini's accept/reject decisions in 95\% of cases, confirming the reliability of automated filtering.


\begin{table}[htbp]
\centering
\small
\caption{Statistics of validated evaluation units by category for random and rule-based augmentation. Categories follow FinNum1~\cite{finnum}.}
\vspace{-0.4cm}
\begin{tabular}{l l r r}
\toprule
\multirow{2}{*}{\textbf{Category}} & \multirow{2}{*}{\textbf{Subcategory}} & \multicolumn{2}{c}{\textbf{Augmentation}} \\
\cmidrule{3-4}
 &  & \textbf{Random} & \textbf{Rule-based} \\
\midrule
\multirow{8}{*}{Monetary} & money & 1,784 & 5,179 \\
         & quote & 1,187 & 2,761 \\
         & change & 1,173 & 3,166 \\
         & forecast & 585 & 1,683 \\
         & buy price & 505 & 1,175 \\
         & support or resistance & 290 & 677 \\
         & sell price & 112 & 262 \\
         & stop loss & 30 & 55 \\
\hline
\multirow{2}{*}{Temporal} & date & 3,635 & 2,033 \\
         & time & 591 & 78 \\
\hline
\multirow{2}{*}{Percentage} & relative & 1,431 & 1,392 \\
           & absolute & 826 & 1,021 \\
\hline
Quantity & quantity & 1,737 & 3,139 \\
\hline
Product Number & product number & 293 & 458 \\
\hline
Indicator & indicator & 272 & 424 \\
\hline
\multirow{2}{*}{Option} & exercise price & 125 & 206 \\
       & maturity date & 75 & 21 \\
\midrule
\textbf{Total} &  & \textbf{14,651} & \textbf{23,730} \\
\bottomrule
\end{tabular}
\label{tab:validated-dataset-stats}
\end{table}

\begin{table*}[h]
\centering
\caption{Main Results.}
\vspace{-0.4cm}
\small
\begin{tabular}{l cc cc cc}
\toprule
 & \multicolumn{2}{c}{\textbf{Triplet (Accuracy)}} & \multicolumn{2}{c}{\textbf{Listwise (Kendall's $\tau_b$)}} & \multicolumn{2}{c}{\textbf{Cross-Pair (Accuracy)}} \\
\cmidrule(lr){2-3}\cmidrule(lr){4-5}\cmidrule(lr){6-7}
\textbf{Checkpoint} & Random & Rule-based & Random & Rule-based & Random & Rule-based \\
\midrule
\texttt{bert-base-uncased}  & 0.9214 & 0.8309 & 0.5409 & 0.3420 & 0.6727 & 0.4815 \\
\texttt{ProsusAI/finbert}   & 0.9186 & 0.8431 & 0.5344 & 0.3580 & 0.6772 & 0.4860\\
\bottomrule
\end{tabular}
\label{tab:main-results}
\end{table*}

\subsection{Evaluation Protocol}\label{sec:evaluation}

To assess metric sensitivity to numerical changes, we employ two evaluation protocols: \textbf{anchor-based evaluation}, which compares augmented variants against their original base sentence,  and \textbf{cross-pair evaluation}, which tests ranking consistency across different sentence contexts. 

\subsubsection{Anchor-based Evaluation}
Anchor-based evaluation tests whe-ther metrics assign higher similarity to numerically closer variants.
Given a base sentence \(s\) with target number $t$ and its augmented variants \(\{\tilde{s}_i\}_{i=1}^{k}\) with modified values  \(\{\tilde{t}_i\}_{i=1}^{k}\), we define numerical distance as
 \(d_\nu(s,\tilde{s_i}) = |\nu(t) - \nu(\tilde{t_i})|\).

\noindent{\bf Triplet evaluation.}
We first identify numerically closest and farthest variants:
\begin{equation} \label{eq:triplet-easy}
\tilde{s}^{+} = \arg\min_i d_\nu(s, \tilde{s}_i), \quad 
\tilde{s}^{-} = \arg\max_i d_\nu(s, \tilde{s}_i).
\end{equation}
A metric passes if it ranks the base sentence as more similar to $\tilde{s}^{+}$ than to $\tilde{s}^{-}$.
We report accuracy over all evaluation units, measuring the proportion of correctly ordered triplets.

\noindent{\bf Listwise evaluation.}
We rank all $k$ variants by their similarity to $s$ and compare this 
predicted ranking against the gold ranking (variants sorted by ascending numerical distance $d_\nu$. 
Larger numerical deviations should yield lower similarity scores.

This helps determine the ability of similarity metrics to perform finer-grained numerical value comparisons across a larger number of sentences. 
We measure ranking quality using Kendall's $\tau_b$ correlation~\cite{kendall-tau}, where $\tau_b=1$ 
indicates perfect agreement, $\tau_b = 0$ indicates no correlation, and $\tau_b = -1$ indicates complete reversal.

\subsubsection{Cross-pair Evaluation}
While anchor-based evaluation tests metrics within a single sentence context, cross-pair evaluation tests whether numerical distance relationships generalize across different sentences. 
This is more challenging because metrics must maintain consistent numerical sensitivity regardless of surrounding text.

We randomly sample two base sentences $s$ and $s^\prime$ from the same numerical category, then independently sample one augmented variant from each to form pairs \((s, \tilde{s}_i)\) and \((s^{\prime}, \tilde{s}^{\prime}_i)\).
The gold ordering is determined by numerical distance: if
\[
d_\nu(s, \tilde{s}_i) < d_\nu(s^{\prime}, \tilde{s}^{\prime}_i),
\] 
then a metric should satisfy $\text{sim}(s, \tilde{s}_i)>\text{sim}(s^\prime, \tilde{s}^\prime_i)$---the pair with smaller numerical deviation should receive higher similarity.
This tests whether metrics preserve correct similarity orderings across different sentence contexts.
We report accuracy, which is the proportion of correctly ordered pairs.

\section{Experiments}

We evaluate BERTScore using two representative checkpoints: \texttt{bert-base-uncased}, the most widely adopted checkpoint in prior work, and \texttt{ProsusAI/finbert}\footnote{\url{https://huggingface.co/ProsusAI/finbert}}, a BERT model further pretrained on 1.8M financial documents including earnings calls, analyst reports, and SEC filings~\cite{araci2019finbert}. 

All experiments use the official BERTScore implementation\footnote{\url{https://github.com/Tiiiger/bert_score}}~\cite{Zhang2020BERTScore} via HuggingFace's \texttt{evaluate}.\footnote{\url{https://huggingface.co/spaces/evaluate-metric/bertscore}} 
We report results for all evaluation protocols introduced in Section~\ref{sec:evaluation}: triplet, cross-pair (accuracy) and listwise (Kendall's $\tau_b$).

\subsection{Results}

Table~\ref{tab:main-results} reveals systematic failures in BERTScore's numerical sensitivity across all evaluation protocols.

\noindent{\bf Triplet evaluation} show the best performance but still exposes fundamental limitations.
Under random augmentation, BERTScore achieves 92.14\% accuracy with bert-base and 91.86\% with FinBERT---meaning it fails to rank the numerically closest variant above the farthest in nearly 8\% of cases, even in this simplified setting. 
Performance degrades substantially under rule-based augmentation, dropping to 83.09\% (bert-base) and 84.31\% (FinBERT). 
This 8–9 percentage point decline reveals difficulty with fine-grained numerical transformations---a critical weakness in finance, where subtle shifts like a one-point change in GDP growth or a 0.1\% difference in profit margins carry major interpretive weight.

\noindent{\bf Listwise evaluation} exposes more severe ranking failures. 
Ken-dall's \(\tau_b\) drops from 0.54 (random) to 0.342  (rule-based) for rule-based for bert-base---barely above weak correlation.
This indicates BERTScore cannot maintain correct similarity orderings when comparing multiple variants with graduated numerical differences, a critical capability for evaluating financial text where precise rankings matter (e.g., identifying the best summary among several with different numerical claims).

\noindent{\bf Cross-pair evaluation} reveals near-complete failure.
Accuracy falls to 48.15\% (bert-base) and 48.6\% (FinBERT) under rule-based augmentation---effectively random performance. 
BERTScore cannot recognize that a 2\% deviation in ``revenue rose 50\%'' should be comparable to a 2\% deviation in ``margins reached 15\%,'' despite identical numerical relationships. This context dependence violates the consistency required for robust evaluation metrics.

FinBERT performs nearly identically to bert-base across all settings (differences <2pp), indicating domain pretraining does not improve numerical sensitivity. These results demonstrate that BERT-Score largely ignores numerical magnitude, treating substantially different values as nearly equivalent—a critical flaw for applications where numerical precision determines meaning.

\subsection{Qualitative Analysis}
To understand BERTScore's numerical failures, consider the tokenization of our motivating example in Section~\ref{sec:motivation}: 
\begin{align*}
s_1 &: [\texttt{revenue}, \texttt{increased}, \texttt{by}, \texttt{3}, \texttt{.}, \texttt{56}, \texttt{\%}, \texttt{.}] \\
s_2 &: [\texttt{revenue}, \texttt{increased}, \texttt{by}, \texttt{4}, \texttt{\%}, \texttt{.}] \\
s_3 &: [\texttt{revenue}, \texttt{increased}, \texttt{by}, \texttt{40}, \texttt{\%}, \texttt{.}]
\end{align*}
Two problems emerge. First, subword tokenizers~\cite{Schuster2012JapaneseAK-wordpiece, sennrich-etal-2016-neural-BPE} fragment numbers into isolated digits, preventing models from learning magnitude-aware representations---``3.56'' becomes three independent tokens with no compositional understanding of the value 3.56.
Second, BERTScore's greedy alignment (Eq.~\ref{eq:token-align}) produces spurious matches: \texttt{4} in $s_2$ aligns to \texttt{40} in $s_3$ based on surface similarity (shared digit), yielding high cosine similarity despite a $10\times$ numerical difference. The sentence-level score thus remains high regardless of magnitude changes.

These issues reveal BERTScore's core limitation: it treats numbers as symbol sequences rather than quantities, explaining its insensitivity to numerical variation in financial text.

\section{Conclusion and Future Work}

We demonstrate that BERTScore fails to capture numerical semantics in financial text, achieving near-random performance (49\% accuracy) when comparing numerically similar sentences across contexts. Using FinNuE, a diagnostic dataset with controlled numerical perturbations, we reveal that BERTScore treats substantially different values (e.g., 2\% vs. 20\%) as nearly equivalent due to subword tokenization fragmenting numbers and greedy alignment ignoring magnitude. Domain pretraining provides no solution: FinBERT performs identically to bert-base. These limitations extend beyond finance to any numerically sensitive domain (scientific literature, medical records, economic data). Future work must develop numerically-aware metrics through: (1) tokenization preserving number boundaries, (2) semantic spaces encoding magnitude relationships, and (3) hybrid approaches combining embeddings with explicit numerical comparison—all feasible without full model retraining. We will release FinNuE publicly to enable the community to benchmark and develop evaluation metrics that faithfully capture numerical semantics.


\bibliographystyle{ACM-Reference-Format}
\bibliography{acm-ref}

\appendix

\section{Prompts Used}\label{app:prompts}

\begin{tcolorbox}[breakable, colback=gray!5, colframe=black, title=Financial Sentence Validation JSON Prompt]
Key Requirements:\par
\vspace{0.2cm}
You are a strict financial text validator.\par
You ONLY output strict JSON, nothing else.\par
Decide if the input sentence is valid or invalid based on numeric/time rules.\par
\vspace{0.2cm}
Input:\par
\textbf{Sentence:} \{sentence\}\par
\vspace{0.2cm}
Rules:\par
- Only treat as invalid in cases like:\par
\hspace{0.4cm}* Invalid time/date format \par
\hspace{0.8cm}Example: \texttt{"\$CMCM at \$10.11 - Buy Stock Market Alert sent to members at 9:4 AM ET \#stocks"}\par
\hspace{0.8cm}$\rightarrow$ reason: \texttt{"Time format '9:4 AM' is invalid; minutes should have two digits (09:04)."}\par
Note: Shortened but still valid expressions like \texttt{"7:00"} (without AM/PM) or \texttt{"11/14"} (without year) are acceptable and should NOT be flagged.\par
\hspace{0.4cm}* Numeric logic inconsistent with the claim \par
\hspace{0.8cm}Example: \texttt{"\$PEIX,is Head-Fake news, all experts point to \$6.02 to \$14 dollars per share that is 130\% upside from here. Art of deal says buy More !"}\par
\hspace{0.8cm}$\rightarrow$ reason: \texttt{"Numeric logic inconsistent: going from \$6.02 to \$14 is more than a 130\% increase."}\par
\vspace{0.2cm}
Output format (strict JSON):\par
\{\par
  \ \ "valid": true/false,\par
  \ \ "reason": "..." \ \# reason required only if invalid\par
\}\par
\end{tcolorbox}

\begin{tcolorbox}[
  breakable,
  colback=gray!5,
  colframe=black,
  title=Number Identification Prompt,
  fontupper=\small,
  left=2mm, right=2mm, top=1mm, bottom=1mm,
  listing only,
  listing options={basicstyle=\ttfamily\footnotesize, breaklines=true}
]
You are a financial text classifier. For each input sentence, identify \textbf{all numeric values} (numbers or percentages) present. For each numeric value, output a separate classification row.\par
Use the FinNum Categories and Subcategories below for classification. The category must start with an uppercase letter. If the subcategory is not the same as the category, start it with a lowercase letter; otherwise, start it with an uppercase letter.\par
\vspace{0.2cm}
Categories and Subcategories:\par
\begin{itemize}
  \item \textbf{Monetary}: money, quote, change, buy price, sell price, forecast, stop loss, support or resistance
  \item \textbf{Percentage}: relative, absolute
  \item \textbf{Option}: exercise price, maturity date
  \item \textbf{Indicator}: Indicator
  \item \textbf{Temporal}: date, time
  \item \textbf{Quantity}: Quantity
  \item \textbf{Product Number}: Product Number
\end{itemize}
\vspace{0.2cm}
Instructions:\par
\begin{itemize}
  \item Output exactly one classification per block with this format:\par
    \begin{verbatim}
Sentence: "<full sentence>"
Target: "<number>"
Category: "<Category>"
Subcategory: "<Subcategory>"
    \end{verbatim}
  \item Output digits only (no symbols, but keep commas).
  \item If no subcategory applies, output "None" for both Category and Subcategory.
  \item Repeat the sentence for each numeric value.
  \item If the sentence contains no numeric values, output a single classification with Target: None, Category and Subcategory both "None".
  \item If there is only a month and year without a specific date, output only the year.
\end{itemize}
\vspace{0.2cm}
Example:\par
\texttt{Sentence: "During 2024, the Company’s net sales through its direct and indirect distribution channels accounted for 38\% and 62\%, respectively, of total net sales."}\\
\texttt{Target: 2024\\ Category: Temporal\\ Subcategory: date}\\

\texttt{Sentence: "During 2024, the Company’s net sales through its direct and indirect distribution channels accounted for 38\% and 62\%, respectively, of total net sales."}\\
\texttt{Target: 38\\ Category: Percentage\\ Subcategory: absolute}\\

\texttt{Sentence: "During 2024, the Company’s net sales through its direct and indirect distribution channels accounted for 38\% and 62\%, respectively, of total net sales."}\\
\texttt{Target: 62\\ Category: Percentage\\ Subcategory: absolute}\\

\texttt{Sentence: "Our common stock has experienced a low of \$138.80 per share."}\\
\texttt{Target: 138.80\\ Category: Monetary\\ Subcategory: quote}

Sentences:

\end{tcolorbox}

\section{Details of Random Augmentation} \label{app:random-augmentation}

\begin{itemize}
  \item \textbf{Temporal}
    \begin{itemize}
      \item \emph{date}: Randomly shift the date within ±30 days, capped by common bounds (e.g., 12, 30, or 60 days).
      \item \emph{time}: 
        \begin{itemize}
          \item If the base is 0: sample between 1 and 60.  
          \item If the base $\leq$ 12: sample between 1 and 12.  
          \item If the base $<$ 60: sample between 1 and 60.  
          \item If the base $\geq$ 60: apply a random shift of about ±20\%.  
        \end{itemize}
    \end{itemize}

  \item \textbf{Monetary}
    \begin{itemize}
      \item \emph{money, quote, forecast, buy/sell price, support or resistance, stop loss}:  
      Randomly increase or decrease the base value, up to ±100\%.
      \item \emph{change}: Apply random shifts within ±100\% of the base.
    \end{itemize}

  \item \textbf{Percentage}
    \begin{itemize}
      \item \emph{relative, absolute}: Randomly increase or decrease the base percentage, up to ±100\%.
    \end{itemize}

  \item \textbf{Quantity}  
    \begin{itemize}
      \item If the base $\leq 5$: allow shifts up to ±300\%.  
      \item Otherwise: apply integer shifts within ±100\% of the base.  
    \end{itemize}

  \item \textbf{Product Number}  
    Randomly increase or decrease the base value, up to ±200\%.

  \item \textbf{Indicator}  
    Randomly increase or decrease the base value, up to ±200\%.

  \item \textbf{Option}
    \begin{itemize}
      \item \emph{maturity date}: Randomly shift the date within ±30 days, capped by common bounds (e.g., 12, 30, or 60 days).  
      \item \emph{exercise price}: Randomly increase or decrease the base value, up to ±200\%.  
    \end{itemize}
\end{itemize}

\end{document}